# COMPARISON OF FORECASTING METHODS OF HOUSE ELECTRICITY CONSUMPTION FOR HONDA SMART HOME

Farshad Ahmadi Asl[1] and Mehmet Bodur[2]

[1]Mathematics and Computer Science Department, Faculty of Arts and Sciences, Eastern Mediterranean University, Famagusta, via Mersin 10, Turkey
[2]Computer Engineering Department, Faculty of Engineering, Eastern Mediterranean University, Famagusta, via Mersin 10, Turkey

## ABSTRACT

*The electricity consumption of buildings composes a major part of the city's energy consumption. Electricity consumption forecasting enables the development of home energy management systems, resulting in the future design of more sustainable houses and a decrease in total energy consumption. Energy performance in buildings is influenced by many factors, like ambient temperature, humidity, and a variety of electrical devices. Therefore, multivariate prediction methods are preferred rather than univariate. The Honda Smart Home US data set was selected to compare three methods for minimizing forecasting errors, MAE and RMSE: Artificial Neural Networks (ANN), Support Vector Regression (SVR), and Fuzzy Rule-Based Systems (FRBS) for Regression by constructing many models for each method on a multivariate data set in different time-terms. The comparison shows that SVR is a superior method over the alternatives.*

## 1. INTRODUCTION

Honda Smart Home was constructed in California, USA, with the goal of creating a sustainable home and a zero-carbon lifestyle [1]. In the domestic energy sector, the development of optimization methods such as *maximum power point tracking* made the use of energy sources like photovoltaic solar energy economically feasible decades ago [2]. Recent studies indicate that the economic optimisation of renewable energy in domestic energy consumption can be further extended by enhancing power management. According to studies, buildings are responsible for the largest proportion of energy consumption in a city, and the residential section is a significant part of it [3][4]. Heating, Ventilating, Air Conditioning (HVAC) systems, and lighting are the main energy-consuming sources of domestic houses. Domestic energy consumption has been increasing due to several factors, like globalisation, greenhouse gas emissions, and population growth [5][6]. For the same reasons, the importance of increasing energy efficiency grows, and electricity forecasting plays a key role in it [7]. For a variety of applications, including management, optimisation, and energy conservation, the importance of accurately forecasting the energy consumption of buildings is emphasized [6]. In addition, accurate energy forecasting models have many implications for the planning and energy optimisation of buildings and are crucial to the economy [8]-[10]. Consequently, energy management utilizing optimized energy forecasting of a domestic house can increase the energy efficiency of the house and the utility of renewable sources such as solar panels.

In the remaining parts of this text, Section 2 explains the time terms and data preparation; Section 3 discusses the details of the models; Section 4 describes the evaluation performance of the models, and Section 5 concludes this work.

## 2. THE HONDA SMART HOME DATA SET AND TIME TERMS

### 2.1. The Honda Smart Home Energy Management System Dataset

In the Honda Smart Home project, which began in 2015 and is ongoing, data related to energy management, such as the HAVC system and Home Energy Management Systems (HEMS), are recorded at a one-minute sampling rate. This forecasting performance study is based on six-month-long energy consumption data from October 2020 to March 2021 of the Honda Smart Home project. The data set is sparse because many of the electric devices work on or off by the resident's decision.

### 2.2. Forecasting Time Terms and Data Preparation

Nonlinear-Multivariate Machine Learning (ML) models for domestic electricity consumption forecasting are built and compared with each other in three-time terms. *Medium-term electricity load forecasting* (MTELF), usually for a week up to a year, which is useful for maintenance scheduling and planning power system outages[11]. *Short-term electricity load forecasting* (STELF), for intervals ranging from one hour to one week; and, one of its primary applications in the daily operation of the electric power system [12]. *Very short-term electricity load forecasting* (VSTELF) ranges from a few minutes to an hour ahead, which is applicable for real-time control, as practiced by [13].

The VSTELF of this study used ten random samples of seventy-minute data collected over a six-month period to get statistical parameters for forecasting performance evaluation. Similarly, the STELF and MTELF used ten pieces of four-day and two-month data randomly selected within the six-month data period for performance comparison of the models, assuming that randomly selected windows reveal the model's weaknesses and strengths better by covering different modes of power consumption.

The original data set contains attributes with one-minute sampling intervals. The VSTELF and STELF followed the original data set's one-minute sampling interval. The MTELF is resampled every ten minutes to reduce the length of the data set to a reasonable size for the forecasting process.

The data set is subdivided as shown in Table 1 in order to evaluate the performance of forecasting independently.

All data sets are prepared in a matrix with eight numerical scalar input variables (attributes) and one scalar target output variable. But occasionally, when a random data set has a column or columns of zero values, some of the forecasting models are unable to scale data to predict the output value. The problem can be solved by omitting the column(s) containing zeros. This circumstance only occurs in VSTELF due to the small size of the data set. Therefore, the input attributes of this time term are variable.

Table 1. Data sets obtained from the Honda database.

| Models | Data sets | Sampling Periods (minute) | Size of Data set | Time Covered | No. input Attributes |
|---|---|---|---|---|---|
| BRNN | VSTELF training + verification | 1 | 30 + 30 | 1 hr. | Variable |
| SVR & SBC | VSTELF training | 1 | 60 | 1 hr. | Variable |
| All models | VSTELF test | 1 | 10 | 10 mins | Variable |
| BRNN | STELF training + verification | 1 | 1500 + 1500 | 2 days | 8 |
| SVR & SBC | STELF training | 1 | 3000 | 2 days | 8 |
| All models | STELF test | 1 | 3000 | 2 days | 8 |
| BRNN | MTELF training | 10 | 2000 + 2000 | 1 month | 8 |
| SVR & SBC | MTELF training | 10 | 4000 | 1 month | 8 |
| All models | MTELF test | 10 | 4000 | 1 month | 8 |

The input variables are the measurement of outdoor temperature and air humidity, as these are significant external factors that affect the house's electricity consumption; and the average power consumption of electric devices (lighting of the living room, lighting of the kitchen, washing machine, refrigerator, microwave, and fans), as these are components of the majority of houses' power consumption. The sum of the average power consumption of the mentioned devices is the target output.

Each attribute in the data matrix has its own units and data range. For successful forecasting, each attribute and output variable are normalized by linearly mapping the columns of the data matrix to the interval [0, 1].

## 3. FORECASTING MODELS

This study tested a number of forecasting models based on the three mentioned methods, which are employed successfully by researchers in regression analysis studies. The best model for each method in terms of performance accuracy was chosen for comparison and evaluation: Bidirectional Recurrent Neural Networks (BRNN) [14]; Support Vector Regression with Analysis of Variance Radial Basis kernel Function (ANOVA RBF) [15][16]; and a combination of the Subtractive Clustering (SBC) method and the Fuzzy C-Means [17][18].

### 3.1. Forecasting by Bidirectional Recurrent Neural Networks

The first model is BRNN, which is computationally expensive compared to basic ANN models such as feed-forward neural networks (FFNN), but the results have shown that the forecasting performance of this model is more accurate for this case study. A basic FFNN and a set of RNNs, including Gated Recurrent Unit (GRU) and Long Short-Term Memory (LSTM), with both ReLU and Tanh activation functions, were tested. The error was calculated by taking the average of 10 runs, and the BRNN with Tanh showed marginally superior accuracy performance with the 10 randomly selected data sets [19]. Figure 1 compares basic FFNN and BRNN models for 2 days of forecasting.

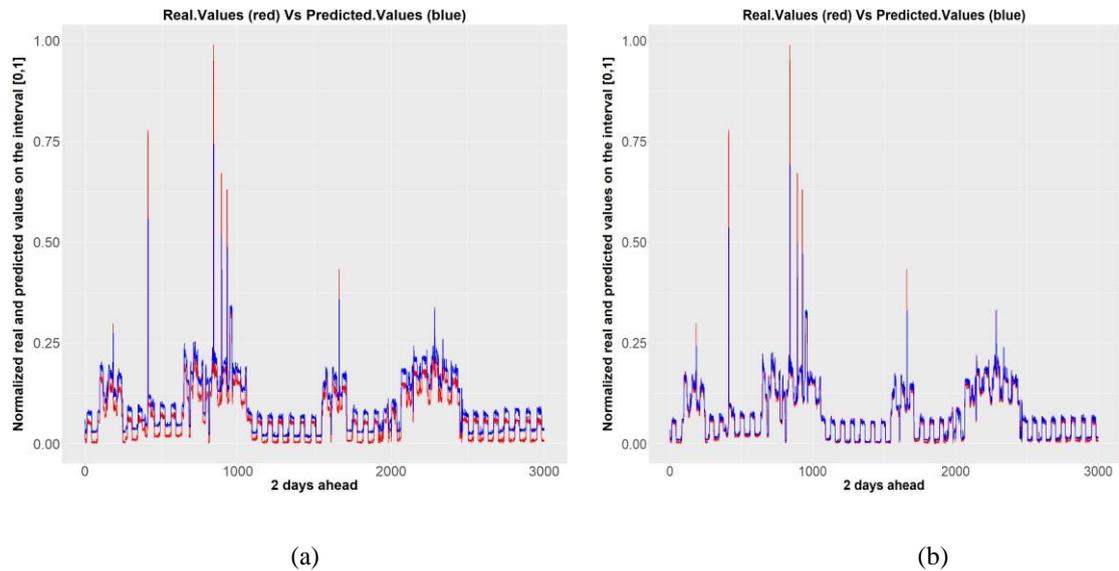

(a)　　　　　　　　　　　　　　　　　　　　(b)

Figure 1. Plots of (a) basic FFNN and (b) BRNN models for 2 days forecasting.

### 3.2. Forecasting by Support Vector Regression Method

The second model, SVR with the ANOVA RBF kernel, is tested from the KERNLAB package [20]. Following a search of various packages and kernel functions such as RBF, Tanh, Bessel, and Laplace, finally, ANOVA RBF from the KERNLAB library package provided outstanding forecasting performance. The SVR method supports regression tasks and employs the Sequential Minimal Optimization (SMO) algorithm. SMO reduces execution time by breaking down the search into multiple sub-search tasks. The SVM saves computational effort by managing the fitting process and the modelling process simultaneously [20]. Figure 2 illustrates the plots of the SVR models with the RBF and ANOVA RBF kernels next to each other from the KERNLAB library package for a better comparison.

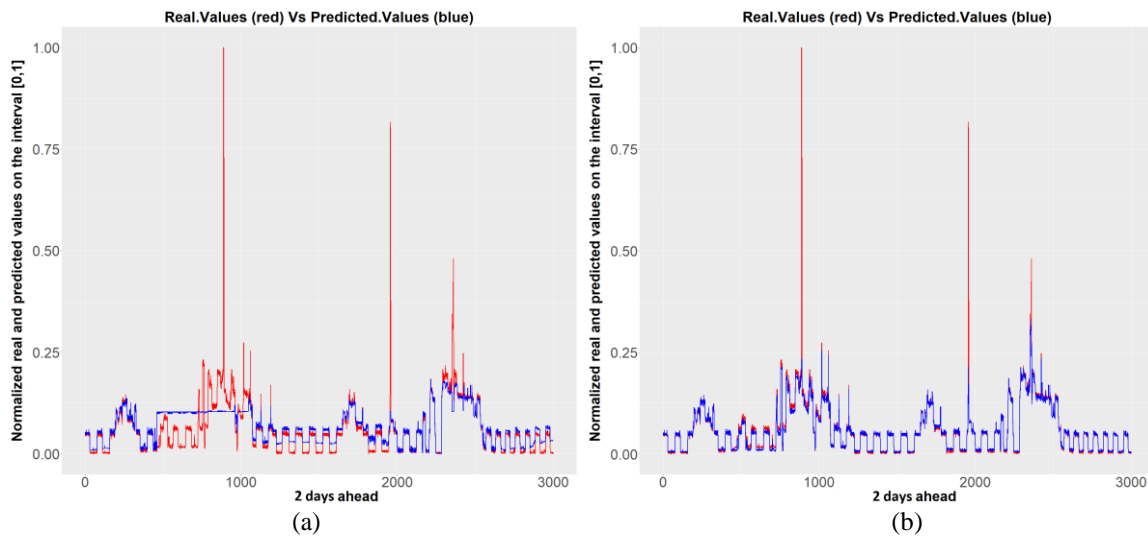

(a)　　　　　　　　　　　　　　　　　　　　(b)

Figure 2. Plots of the SVR with the (a) RBF and (b) ANOVA RBF Kernels.

### 3.3. Forecasting by Fuzzy Rule Base System with Subtractive Clustering

The third model is a combination of the SBC method and the Fuzzy C-Means technique from the FRBS package [22]. SBC considers each data point as a potential cluster centre and calculates the likelihood of each data point defining a cluster centre based on its distance to all other data points. The point with the highest potential among the remaining points is chosen as the next cluster centre. Afterward, the process repeats until all cluster centres are obtained. The Fuzzy C-Means algorithm is then used to optimize the cluster centres [17].

A set of models such as Fuzzy Rule-Based Systems based on space partition, neural networks, clustering approach, and the gradient descent method were evaluated from the FRBS package to determine the model with the highest performance accuracy. The majority of them resulted in poor forecasting performance and long runtimes, which made them inadequate for forecasting very short-term electricity consumption. Consequently, the SBC model has been considered acceptable for power consumption forecasting. Side-by-side comparisons of the Hybrid Neural Fuzzy Inference System (HyFIS) and the SBC model with C-mean optimization are presented in Figure 3. Both models are available in the FRBS package.

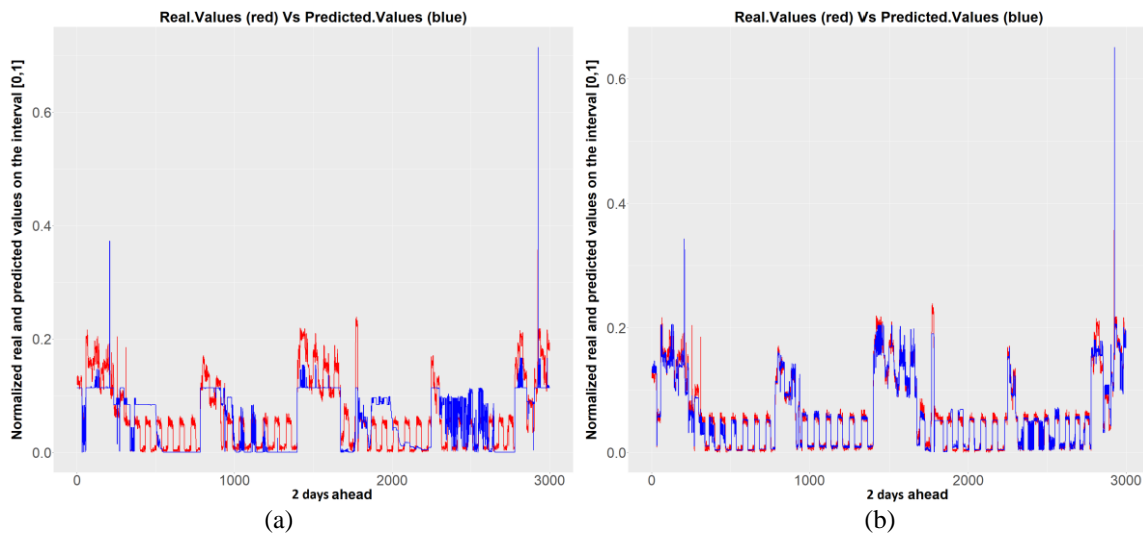

Figure 3. Plots of the models, (a) HyFIS and (b) SBC.

All of the written codes for the models in the R programming language can be found in [23].

### 4. MODEL PERFORMANCE EVALUATION

The accuracy of all models is measured in two metrics: Mean Absolute Error (MAE) and Root Mean Square Error (RMSE). These two metrics are used in this work as they are the most common for measuring the accuracy of electricity forecasting and thus make this study more comparable to the others. The models are comparable since they use the same test data and normalization method ([0–1] min–max normalization). Additionally, the models' execution times are measured. Table 2 contains the average MAE and RMSE values of 10 repeated runs of each model for each time term. The average execution time of the 10 repeated runs of each model for each time term is given in Table 3.

Table 2. Evaluation of the models: Average of 10 runs with 10 random data samples.

| Time-Terms Models | VSTELF | | STELF | | MTELF | |
|---|---|---|---|---|---|---|
| | MAE | RMSE | MAE | RMSE | MAE | RMSE |
| SVR | 0.009 | 0.013 | 0.005 | 0.019 | 0.0038 | 0.014 |
| SBC | 0.027 | 0.042 | 0.01 | 0.027 | 0.0055 | 0.014 |
| BRNN | 0.11 | 0.126 | 0.011 | 0.031 | 0.024 | 0.046 |

Table 3. Execution Time of the models: Average of 10 runs with 10 random data samples.

| Time-Terms Models | VSTELF Execution time | STELF Execution time | MTELF Execution time |
|---|---|---|---|
| SVR | less than 10 sec | less than 10 sec | less than 10 sec |
| SBC | less than 10 sec | Around 5 min | Around 9 min |
| BRNN | Around 15 sec | Around 30 sec | Around 45 sec |

As demonstrated in Figure 4, the BRNN and SBC models are not ideal for very short-term forecasting. In contrast, the SVR model is suitable for VSTELF with a decent result. The execution times of the models are around 10 seconds in this time term.

The performances of the SBC and BRNN models for short-term forecasting are fairly similar. The results showed that these two models are more accurate with a larger data training set, whereas the SBC model's execution time becomes noticeably longer as the training set grows. The performance and execution time of the SVR model compared to the other two models are better in this time term. Model plots for side-by-side comparison are shown in Figure 5.

The BRNN model performs marginally worse than the SBC for medium-term forecasting. It can be the result of data set sampling. On the contrary, SBC showed its best forecasting performance. SVR is the model with the most accurate forecasting performance among the others. The execution times of the models are nearly identical to STELF, with the exception of SBC, which requires approximately four minutes longer to complete. Figure 6 compares the plots of the models in this time term.

Similar to the research that showing STELF is a suitable area for the implementation of neural networks, the BRNN model in this study showed its best performance in this time term [24]. The SBC model demonstrated decent performance in terms of performance accuracy while working with a large data training set, but it makes it inadequate for VSTELF. Also, it was observed that the SBC model could not predict all of the output values and express them as undefined values.

Both SVR and SBC models were unable to scale the model to predict the output value when a random data training set contained zero values in one or more columns. The issue was solved by removing the zero column/columns from the data.

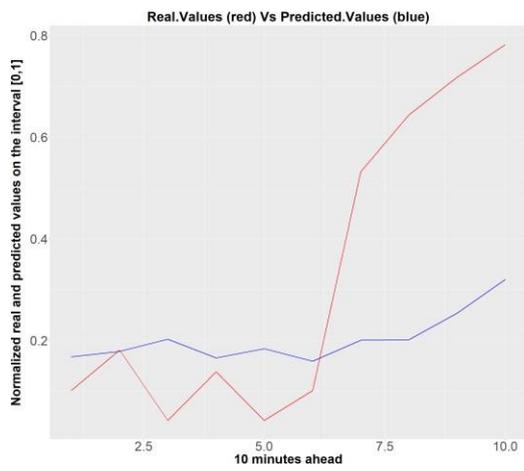

(a)

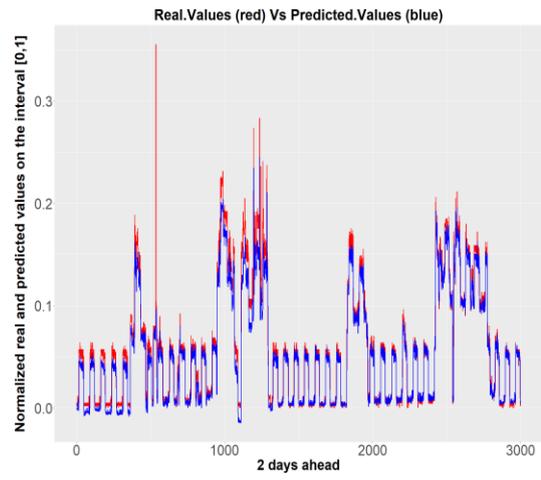

(a)

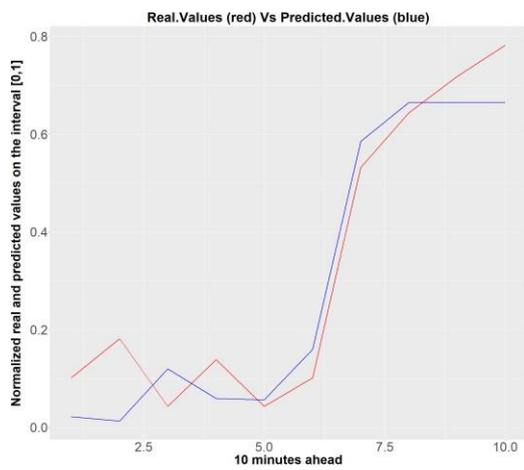

(a)     (b)

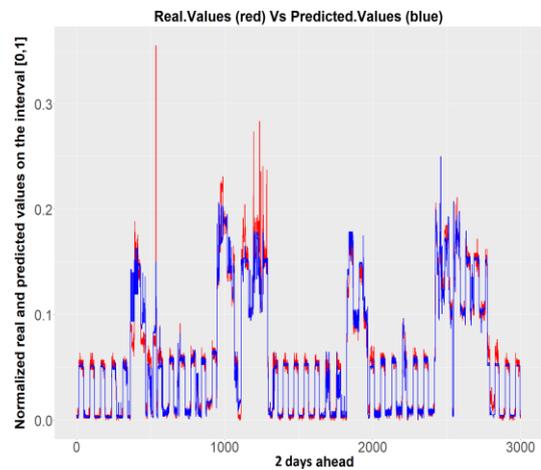

(b)

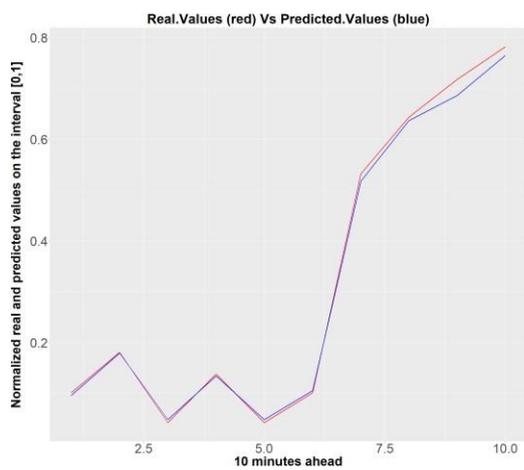

(b)     (c)

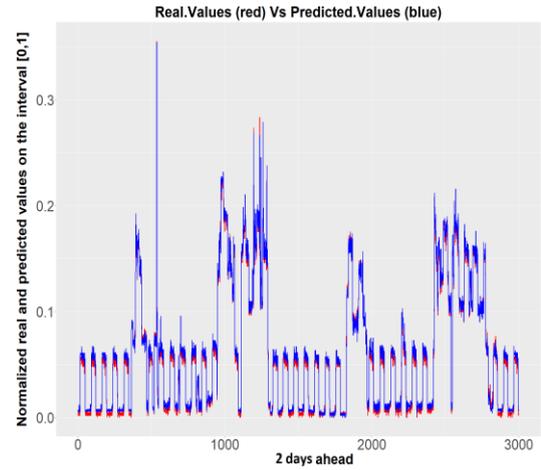

(c)

Figure 4. Plots for (a) BRNN model, (b) SBC model, (c) SVR model in VSTELF.

Figure 5. Plots for (a) BRNN model, (b) SBC model, (c) SVR model in STELF.

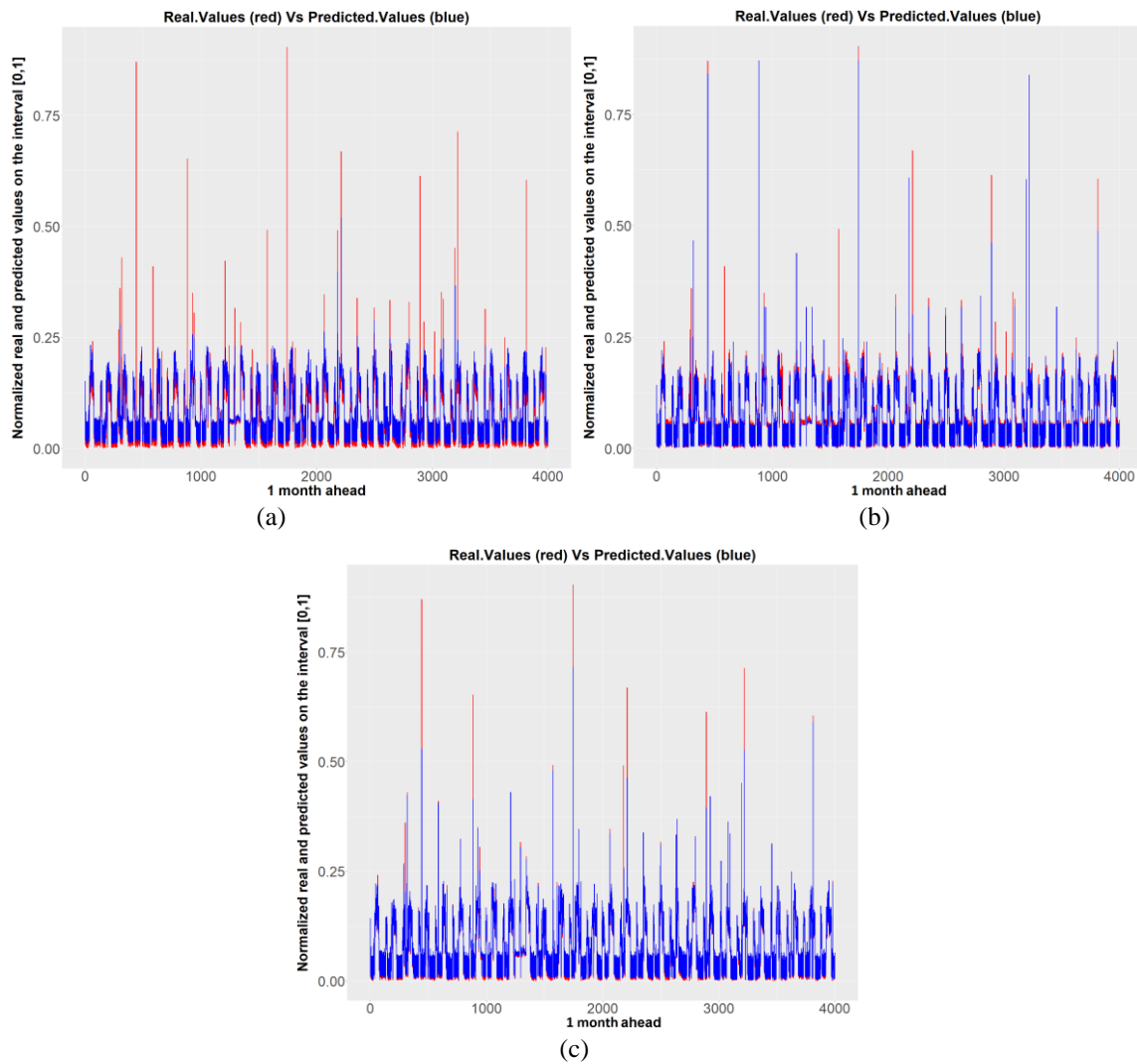

Figure 6. Plots for (a) BRNN model, (b) SBC model, (c) SVR model in MTELF.

## 5. CONCLUSIONS

This study compared the performance of three widely-used methods for forecasting the electricity consumption of domestic houses over three forecasting time terms. The strengths and weaknesses of each method were observed across different data set sizes, time terms, and execution times. In addition, the best model constructed using a single method is identified for this case study.

The BRNN model uses a bidirectional layer that processes a sequence in both directions, making the model ideal for time series forecasting [19]. This is one of the reasons why this model is more accurate than the other RNN models tested. In addition, the Tanh activation function activates almost all the input neurons to predict the output, which makes it more computationally expensive but more accurate than the other activation functions, thereby enhancing the model's performance accuracy. Nevertheless, this model is suitable for STELF and MTELF but not ideal for VSTELF.

The SVR model was constructed using the KERNLAB library package and employs an SMO optimization algorithm during the modelling process, which is significantly faster than data

Computer Science & Information Technology (CS & IT)                                    139

deduplication techniques such as the chunking algorithm on sparse data sets [21]. ANOVA, by analysing and comparing differences between group means or population means (of variables) and their associated procedures, such as variation [25][26], helps the RBF kernel and the model for a precise forecast. This model handles both the fitting and the modelling processes at the same time, saving computational effort and making it suitable for forecasting in all time terms.

In terms of forecasting accuracy, the SBC model with the Fuzzy C-mean optimization outperformed the other tested models in the FRBS library package. However, it is inefficient for very short-term forecasting, especially with a large training data set, but it is suitable for STELF and MTELF if a long execution time is not a concern for the forecast.

The results indicate that the selected SVR model forecasts with a lower mean absolute and root mean square error than the other models in all time terms. Additionally, this model is suitable for very short-term forecasting since its execution time is fast, even for large data sets. Moreover, the simple implementation of the SVR model makes it an excellent choice for forecasting in all time terms for time series data.